\pdfoutput=1
\documentclass[conference]{IEEEtran}
\IEEEoverridecommandlockouts
\usepackage{cite}
\usepackage{amsmath,amssymb,amsfonts}
\usepackage{algorithmic}
\usepackage{graphicx}
\usepackage{booktabs} 
\usepackage{longtable}
\usepackage{supertabular}
\usepackage{easyReview}
\usepackage{listings}
\usepackage{textcomp}
\usepackage{xcolor}
\usepackage{pythonhighlight}
\usepackage{multicol}
\usepackage{tabularray}  
\usepackage{amssymb}
\usepackage{lipsum}
\makeatletter
\newcommand{\linebreakand}{%
  \end{@IEEEauthorhalign}
  \hfill\mbox{}\par
  \mbox{}\hfill\begin{@IEEEauthorhalign}
}
\makeatother

\def\BibTeX{{\rm B\kern-.05em{\sc i\kern-.025em b}\kern-.08em
    T\kern-.1667em\lower.7ex\hbox{E}\kern-.125emX}}
\begin{document}

\title{Autocompletion of Chief Complaints in the Electronic Health Records using Large Language Models\\
\thanks{This work is funded by Northwestern Mutual Data Science Institute (NMDSI), Milwaukee, WI, USA.}
}

\author{
    \IEEEauthorblockN{1\textsuperscript{st} K M Sajjadul Islam}
    \IEEEauthorblockA{\textit{Computer Science} \\
    \textit{Marquette University}\\
    sajjad.islam@marquette.edu}
\and
    \IEEEauthorblockN{2\textsuperscript{nd} Ayesha Siddika Nipu}
    \IEEEauthorblockA{\textit{Computer Science \& Software Engineering} \\
    \textit{University of Wisconsin-Platteville}\\
    nipua@uwplatt.edu}
\linebreakand
    \IEEEauthorblockN{3\textsuperscript{rd} Praveen Madiraju}
    \IEEEauthorblockA{\textit{Computer Science} \\
    \textit{Marquette University}\\
    praveen.madiraju@marquette.edu}
\and
    \IEEEauthorblockN{4\textsuperscript{th} Priya Deshpande}
    \IEEEauthorblockA{\textit{Electrical \& Computer Engineering} \\
    \textit{Marquette University}\\
    priya.deshpande@marquette.edu}
}

\maketitle 

\begin{abstract}
The Chief Complaint (CC) is a crucial component of a patient\textquotesingle s medical record as it describes the main reason or concern for seeking medical care. It provides critical information for healthcare providers to make informed decisions about patient care. However, documenting CCs can be time-consuming for healthcare providers, especially in busy emergency departments. 
To address this issue, an autocompletion tool that suggests accurate and well-formatted phrases or sentences for clinical notes can be a valuable resource for triage nurses. In this study, we utilized text generation techniques to develop machine learning models using CC data. 
In our proposed work, we train a Long Short-Term Memory (LSTM) model and fine-tune three different variants of Biomedical Generative Pretrained Transformers (BioGPT), namely microsoft/biogpt, microsoft/BioGPT-Large, and microsoft/BioGPT-Large-PubMedQA. Additionally, we tune a prompt by incorporating exemplar CC sentences, utilizing the OpenAI API of GPT-4.
We evaluate the models\textquotesingle \space performance based on the perplexity score, modified BERTScore, and cosine similarity score. 
The results show that BioGPT-Large exhibits superior performance compared to the other models. It consistently achieves a remarkably low perplexity score of 1.65 when generating CC, whereas the baseline LSTM model achieves the best perplexity score of 170. Further, we evaluate and assess the proposed models\textquotesingle \space performance and the outcome of GPT-4.0.
Our study demonstrates that utilizing LLMs such as BioGPT, leads to the development of an effective autocompletion tool for generating CC documentation in healthcare settings.

\end{abstract}


\section{Introduction}
A chief complaint (CC) is a brief statement that explains why a patient is seeing a doctor. It is usually the second thing asked during a medical history after identifying the patient\textquotesingle s demographic information \cite{chang2019generating}. When a patient seeks medical care, their CC is recorded several times. First, when they register at a clinic or emergency department (ED), triage nurses and clerks create a record. Then, clinicians also document the CC in various notes throughout the patient\textquotesingle s care, including daily progress notes, discharge notes, transfer notes, and patient acceptance summary notes \cite{wagner2006chief}. The limited time and information available during triage can sometimes result in an oversimplified or inaccurate CC, which may not fully capture the patient\textquotesingle s symptoms or concerns. This can potentially impact the diagnostic process, as the treating clinician may not have a complete understanding of the patient\textquotesingle s condition and may not order appropriate tests or treatments \cite{Misleading_Complaint:online}. In addition, errors in CC\textquotesingle s can also occur due to misspelled words, incorrect punctuation, or inaccurate symptom descriptions \cite{karagounis2020coding}.

The goal of this study is to employ Natural Language Processing (NLP) techniques to create an autocompletion tool for CC\textquotesingle s in ED settings. A state-of-the-art (SOTA) NLP model may help triage nurses generate accurate CC\textquotesingle s more efficiently. This study aims to 

\begin{itemize}

\item Explore the potential of NLP techniques for autocompleting CC\textquotesingle s in ED settings. This study will involve developing an NLP model capable of generating CC\textquotesingle s. This generated CC will not only suggest accurate and well-formatted notes but also provide ideas to improve their notes.

\item Assess the impact of an autocompletion tool on the efficiency and accuracy of triage in ED settings. This study will compare the accuracy of CC\textquotesingle s generated with the NLP model to those entered manually by triage healthcare providers. 

\end{itemize}

Autocompletion provides word, phrase, or sentence suggestions as a user types. The primary objective of this system is to improve efficiency by reducing the number of keystrokes required, while also elevating the quality of the content by minimizing typographical errors, promoting the adoption of standardized terminology, and facilitating the exploration of a wider range of vocabulary \cite{spithourakis2016clinical}. This process works by analyzing previously entered words to make educated guesses about a subsequent word, phrase, or sentence. To complete a CC automatically, text generation techniques are employed which is one of the primary tasks in Natural Language Generation (NLG). NLG is a specialized area within the discipline of NLP that focuses on the development of systems with the ability to generate both coherent and easily understandable text. NLG is often regarded as a comprehensive term that incorporates a diverse array of tasks involving the transformation of input data into a textual sequence as output. These tasks include generating answers for users in a chatbot, translating languages, suggesting story ideas, or summarizing data analysis. Clinical documents provide distinct issues in comparison to general-domain text due to the extensive utilization of acronyms and non-standard clinical terminology by healthcare professionals, as well as the irregular structure and arrangement of these documents \cite{hasan2019clinical}. Although Generative Pretrained Transformers (GPT) models \cite{radford2018improving, radford2019language,brown2020language} demonstrate proficiency in generating coherent text for broad subject areas, their effectiveness may diminish when confronted with the complexities inherent in clinical documentation. 

CC\textquotesingle s are free text that consists of one or more improper sentences and medical acronyms \cite{osborne2020gout}. General-purpose language models may not be able to capture the context and fail to show exemplary results on CC\textquotesingle s. GPT-2 \cite{radford2019language} has recently adapted to the bio-medical domain. Biomedical Generative Pretrained Transformers (BioGPT) is such an adaptation that has been trained on a very large corpus of biomedical literature and has shown to work well on many tasks, including text generation \cite{luo2022biogpt}. Hence, we propose to employ BioGPT for autocompletion of the CC.

\section{Background study}
\subsection{Chief Complaint}
The ED in hospitals gets very crowded; it often has more patients and fewer resources than other departments. Many studies show that when the ED is too crowded, the quality of care for patients gets worse \cite{tiwari2014arrival}. 
Long wait time at the point of triage in ED causes patient dissatisfaction \cite{shah2015managing}. 
Patients may have to wait a long time for treatment or to leave the ED. Overcrowding can also lead to medical errors and bad outcomes for patients \cite{kulstad2010ed}. 
The Emergency Nurses Association (ENA) Triage curriculum stresses the significance of CC in the decision-making process for emergency nurses \cite{travers2003using}. It is the first piece of information gathered during the triage assessment.
Around 20\% of patients who visit an ED have non-specific complaints and the majority of them are elderly. Research conducted by retrospective chart analysis indicates that these patients are at a higher risk of being misdiagnosed and require hospital admission \cite{sauter2018non}. 
A study conducted by Nunez et al. (2006) demonstrated that the lack of seriousness of the initial CC is a major factor in patients\textquotesingle \space unscheduled return to ED \cite{nunez2006unscheduled}. An autocompletion tool for CC can help alleviate these problems.

Several studies have been done with CC datasets.
Tootooni et al. (2019) proposed a heuristic methodology for automatically mapping free-text CC data into a structured list of CCs, using an NLP-based algorithm called Chief Complaint Mapper (CCMapper) and to demonstrate its high performance and capability of incorporating new free-text CC data \cite{tootooni2019ccmapper}.
Chang et al. (2020) used the Bidirectional Encoder Representations from Transformers (BERT) language model to learn contextual embeddings for CC \cite{chang2020generating}. It predicts their provider-assigned labels with potential applications in automating the mapping of free-text CC\textquotesingle s to structured fields and developing a standardized ontology. 
Hsu et al. (2020) used NLP technologies, including deep learning methods such as BERT, to classify Chinese CC\textquotesingle s at emergency departments for the detection of influenza-like illness, with the goal of developing a fast and effective tool to assist physicians in making diagnoses and controlling outbreaks \cite{hsu2020natural}.

\subsection{Text Generation in Electronic Health Record}
The process of generating Electronic Health Records (EHRs) presents significant challenges due to the complex diverse nature of medical data, the imperative for utmost accuracy, and the rigorous demands for privacy. 
Recent improvement in NLG is revolutionizing EHR generation in different fields such as report generation from medical images \cite{jing2017automatic}, medical note generation from table data \cite{wu2022medical}, medical topic to text generation \cite{pan2020medwriter}, and so on. More focus has been given to synthetic EHR generation due to the scarcity of medical data \cite{guan2018generation, melamud2019towards, amin2020exploring, tang2023does}. 
In their work, Lee et al. (2018) generate synthetic CC\textquotesingle s from discrete variables in EHRs, like age group, gender, and discharge diagnosis \cite{lee2018natural}.  

Recent advancements in EHR generation have leveraged a range of methodologies, from Long Short-Term Memory (LSTM) to transformer-based language modeling. 
In a study by Liu et al. (2018), a novel transformer-based language modeling job was introduced. This work involved predicting the content of medical notes, taking into consideration previous data from a patient\textquotesingle s medical record \cite{liu2018learning}. 
Krishna et al. (2020)  primarily used LSTM and BERT to generate semi-structured clinical summaries (SOAP) notes from doctor-patient conversations \cite{krishna2020generating}. 
Ive et al. (2020) used a neural Transformer model to generate artificial clinical documents for mental health records \cite{ive2020generation}. 
Sirrianni et al. (2022) employed GPT-2 and GPT-Neo for next-word prediction on dental medical notes that include exam notes, emergency notes, trauma notes, etc \cite{sirrianni2022medical}.

\subsection{Autocompletion in Electronic Health Record}
Over the past few years, researchers have  extensively investigated diverse techniques to enhance autocompletion tasks in the medical domain. Spithourakis et al. (2016) developed LSTM-based neural language models to improve word prediction and completion tasks \cite{spithourakis2016clinical}. They demonstrated superior performance on a clinical dataset. 
Yazdani et al. (2019) investigated the effectiveness of a tri-gram language model in predicting the next words while typing free texts \cite{yazdani2019words}. 
Van et al. (2020) explored the use of autocomplete and pre-trained neural language models in semi-automated text simplification for the medical domain, using a new parallel dataset, and comparing the performance of four models and an ensemble model \cite{van2020automets}.

In the biomedical domain, the scarcity of large-scale annotated data makes it essential to use pre-trained language models, which can act as rich feature extractors and reduce reliance on annotated samples \cite{wang2021pre}. These models also serve as soft knowledge bases, capturing the domain\textquotesingle s intricate knowledge from vast unannotated texts. In the biomedical field, there has been a significant increase in the attention given to PLM models such as clinical BERT and BioGPT in recent years.
To the best of our knowledge, we have not come across any research that specifically addresses autocompletion using SOTA biomedical-based PLMs for CC datasets.


\begin{figure}[tbp]
\centerline{\includegraphics[width=0.48\textwidth]{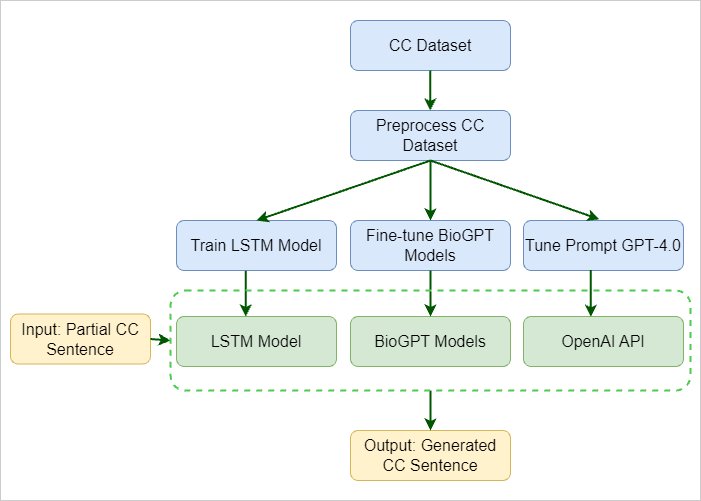}}
\caption{Process Flow of Current Study}
\label{Fig:ProcessFlow}
\end{figure}

\section{Methodology}

Text generation has evolved significantly from its early days of statistical language models to neural networks.
Jozefowicz et al. (2016) showed that training recurrent neural network (RNN) LMs on extensive datasets yields superior performance compared to other statistical language models, such as meticulously optimized N-grams \cite{jozefowicz2016exploring}. While neural models have made impressive advancements in text generation, their performance is often hindered by the scarcity of expensive labeled data \cite{sutskever2014sequence}. 
However, the inception of the Transformer architecture\cite{vaswani2017attention}, which is the foundation of pre-trained language models, marked a significant advancement. Pre-trained models have revolutionized the capabilities of text generation exhibiting improved accuracy and fluency. There exist two primary categories of pre-training models: BERT-like models \cite{devlin2018bert, lan2019albert, alsentzer2019publicly} are primarily utilized for language understanding tasks, while the GPT-like models \cite{radford2018improving, luo2022biogpt} are primarily employed for language generation tasks.

Our study suggests that LSTM and BioGPT, are the most suitable models for our tasks. LSTM model is widely recognized and commonly employed as a baseline \cite{melis2017state} and BioGPT demonstrates impressive capabilities in NLG, especially in the medical domain \cite{luo2022biogpt}. Additionally, OpenAI API \cite{OpenAI_API:online} from the GPT-4.0 model, is utilized to develop a prompt by implementing few-shot (FS) technique. Figure~\ref{Fig:ProcessFlow} depicts the overall flow of our study. 

\subsection{Dataset Description}\label{Sec:DatasetDescription}
Osborne et al. (2020) developed an algorithm for identifying gout flares in ED patients using triage nurse CC notes \cite{osborne2020gout}. In this work, the researchers have provided a de-identified version of a clinical corpus which to the best of their knowledge, is the first free-text CC clinical corpus available. 
The corpus was de-identified to adhere to Health Insurance Portability and Accountability Act (HIPAA) Safe Harbor regulations. This de-identification process involves fine-tuning named entity recognition algorithms using BERT \cite{devlin2018bert} and ALBERT \cite{lan2019albert}. In addition, potentially identifiable time information was eliminated, followed by a thorough manual review utilizing BRAT software\cite{stenetorp2012brat} to guarantee the absence of personal information. 
The corpus was annotated to predict gout flare status based on a retrospective manual examination of CC\textquotesingle s. A subset of these complaints underwent review by rheumatologists, applying Gaffo criteria to confirm gout flare status, with annotator agreement calculated for both the initial annotation and chart review phases.
This publicly available corpus consists of 2 datasets: GOUT-CC-2019-CORPUS and GOUT-CC-2020-CORPUS. In the corpus, there are in total of 8342 CC and each observation has 3 fields: CC, predict, and consensus. 
The ``Chief Complaint" field consists of freely written text with abundant abbreviations and acronyms. The ``Predict" field signifies potential gout flare relevance (Y, N, U, -), while the ``Consensus" field indicates gout flare status based on chart review (Y, N, U, -). Here values are yes (Y), no (N), unknown (U), or unmarked (-).
For our purpose, we only employed CC data. The first 3 observations from the dataset are mentioned in Table~\ref{dataset-table}.

\begin{table}[tbp]
\caption{Sample of Chief Complaint Dataset}
\begin{center}
\begin{tabular}{p{5cm}ll}
\hline
\textbf{Chief Complaint $^{\mathrm{a}}$} & \textbf{\textit{Predict}}& \textbf{\textit{Consensus}}\\
\hline``been feeling bad" last 2 weeks \& switched BP medications last week \& worried about BP PMHx: CHF, HTN, gout, 3 strokes, DM & N & - \\
\hline
``can\textquotesingle t walk", reports onset at \textless \textless TIME\textgreater \textgreater. oriented x2. aortic valve replacement in \textless \textless DATE \textgreater \textgreater. wife reports episode of similar last week, hospitalized at \textless \textless HOSPITAL\textgreater \textgreater for UTI, gout - pmhx: CVA (L side residual deficits) & Y & N \\
\hline
``dehydration" Chest hurts, hips hurt, cramps PMH- Hip replacement, gout, missed pain clinic appt today, thinks he has a gout flair up knee and foot pain & Y & Y \\
\hline
\multicolumn{3}{l}{$^{\mathrm{a}}$Only CC column is employed in present work.}
\end{tabular}
\label{dataset-table}
\end{center}
\end{table}

\subsection{Data Preprocessing}\label{sec_data_preprocess}
\begin{figure}[tbp]
\centerline{\includegraphics[width=0.45\textwidth]{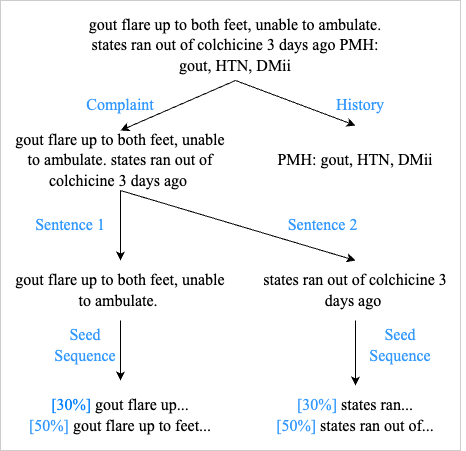}}
\caption{Illustration of Preprocessing Steps with Example}
\label{data_preprocess_example}
\end{figure}

CC is a free text which consists of one or more improper sentences. It is mostly written in abbreviated forms and enriched in medical acronyms. From our observation, we identify that a CC consists of 2 parts, the first part involves a patient\textquotesingle s complaint regarding their current health condition, and the second part pertains to their past medical or personal history. We find several medical acronyms that describe past medical or personal history such as PMH, PMHX, HX, PSHX, SHX, and FHX. We split a CC into two parts based on past medical or personal history. The complaint part consists of one or more improper sentences. We use the Python NLP library Stanza to separate sentences. After splitting each CC in sentences, we filter them based on the length. If a sentence contains less than 4 words, it is discarded from the dataset. For instance: ‘Denies nausea’, ‘24 weeks OB’, etc. are filtered from further consideration as these types of small sentences do not require autocompletion and degrade model performance. We find a total of 11770 sentences after splitting CC and filtering the small sentences. The dataset is divided into three sets - train, validation, and test; with a ratio of 80\%, 10\%, and 10\%, respectively. The vocabulary size in the training set is 11565 and the median number of words per sentence is 9 which indicates a higher level of diversity in the dataset. It is expected that the user will type 3 or 4 words initially which is 30\% to 50\% of the sentence. For every test sentence, 2 seed sequences are generated by taking 30\% and 50\% from the beginning. A data preprocessing example is shown in Figure~\ref{data_preprocess_example}.


\begin{figure}[tbp]
\centerline{\includegraphics[width=0.427\textwidth]{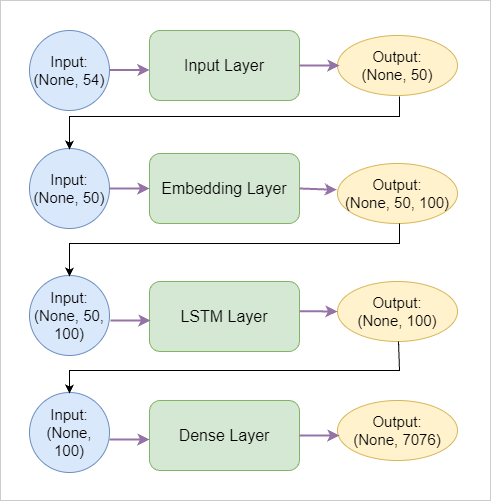}}
\caption{Framework of Proposed LSTM Model Architecture}
\label{LSTM_model_arc}
\end{figure}

\subsection{A Neural Network Approach}\label{Sec-LSTM}
LSTM \cite{hochreiter1997long} is a type of RNN that has shown high-quality performance in NLP tasks \cite{park2018word, yao2018improved}. RNNs are specifically engineered to effectively process sequential input by employing a hidden state that undergoes iterative updates at each consecutive step. LSTM networks possess unique gating mechanisms, enabling them to effectively capture and learn long-term dependencies. In LSTM, the model can selectively choose which information to keep or forget from the previous state, making it more capable of handling long-term dependencies in the input data \cite{wen2015semantically}. The ability of LSTMs to effectively handle sequential input and comprehend long-term contextual information serves as a foundation of text generation, which is the iterative process of making predictions for the subsequent word in a sequence.

Figure~\ref{LSTM_model_arc} illustrates our proposed LSTM model for text generation. The first layer of the model is an Embedding layer which is used to convert the input text data into dense word vectors of 100 dimensions. This layer takes three arguments: the total number of unique words in the input corpus, the dimensionality of the embedding space, and the maximum length of input sequences. The next layer is an LSTM layer with 100 LSTM cells, a type of RNN layer that processes input data to capture long-term dependencies in the text. The output of the LSTM layer is then passed to a Dense layer with the number of neurons and softmax activation function. This layer generates the probability distribution of the next word in the sequence, given the input sequence. We enable Adam optimizer, a popular optimizer used for gradient descent in deep learning, with a learning rate of 0.001. We utilize categorical cross-entropy loss function that is widely used for multi-class classification tasks.  
This model is capable of predicting subsequent words in a sequence. During training, each sentence is prepended with an \textless sos\textgreater \space token and appended with an \textless eos\textgreater \space token to signify the start and end. However, the model struggles to accurately identify sentence endings in its predictions. As a workaround, we apply an iterative approach to generate the next five words in any given sequence, regardless of sentence boundaries.

\subsection{A Transfer Learning Approach}\label{AA}
BioGPT is a highly specialized generative pre-trained Transformer language model that has been specifically designed and optimized for the purpose of generating and analyzing biomedical texts \cite{luo2022biogpt}. The model architecture was derived from the GPT-2 \cite{radford2019language} model architecture and serves as its backbone. Its training process involves utilizing a dataset including 15 million abstracts sourced from PubMed. The ultimate acquired vocabulary size amounts to 42,384. The GPT-2 (medium) model, serving as the foundation network, consists of 24 layers, a hidden size of 1024, and 16 attention heads. This configuration yields a total of 355 million parameters. On the other hand, the BioGPT model has 347 million parameters. The difference comes solely from variations in the embedding size and output projection size, which are a consequence of the dissimilar vocabulary sizes. BioGPT also scaled to larger size. The BioGPT- Large model was built with the GPT-2 XL architecture, which represents the most extensive iteration of GPT-2, having a total of 1.5 billion model parameters. The BioGPT models demonstrate exceptional performance on four benchmark datasets, namely BC5CDR, KD-DTI, DDI end-to-end relation extraction job, and PubMedQA question answering test, surpassing previous SOTA approaches. In addition, the model depicts better biomedical text-generation proficiency in comparison to a standard GPT model trained on a general domain.

 Pretrained BioGPT can be adapted from downstream tasks such as end-to-end relation extraction, question answering (QA), and document classification by fine-tuning the model.  For this work, we tailor the model specifically for text generation. To fine-tune BioGPT, we utilize Raj-High Performance Computer which is funded in part by the National Science Foundation award CNS-1828649 ``MRI: Acquisition of iMARC: High Performance Computing for STEM Research and Education in Southeast Wisconsin" \cite{Raj:online}.  

Pretrained BioGPT models are available in Huggingface directory. For text generation, we fine-tune `microsoft/biogpt'\footnote{https://huggingface.co/microsoft/biogpt}, `microsoft/BioGPT-Large'\footnote{https://huggingface.co/microsoft/BioGPT-Large} and `microsoft/BioGPT-Large-PubMedQA'\footnote{https://huggingface.co/microsoft/BioGPT-Large-PubMedQA}. We exploit the tokenizer from the same models and tokenize the input sequences by adding special tokens \textless sos\textgreater \space (start of sentence) and \textless eos\textgreater \space (end of sentence) at the beginning and end of each sentence, respectively. Subsequently, padding is performed considering the maximum token sequence (74 tokens) to make the dimension uniform regardless of the input sequence. Additionally, Adam optimizer is incorporated into the model\textquotesingle s training pipeline.  

BioGPT models possess the capacity to generate multiple sequences for a single seed sequence. For each of the seed sequences, we assign the number of return sequences to 5. The `generate' function from huggingface includes additional options such as do\_sample, top\_k, max\_length, top\_p, etc., which serve to regulate the output sequence. The boolean flag do\_sample is utilized to decide whether or not to employ sampling throughout the process of text generation. The parameter top\_k is an integer that determines the number of most probable words to be taken into account while generating text. The variable max\_length is an integer that serves as a control parameter for determining the maximum length of the output text. The variable top\_p is a floating-point number that determines the cumulative probability of selecting the most frequent words to be considered in the process of generating text.

\subsection{Prompt Tuning: Few-Shot Technique}\label{AAA}

OpenAI provides API to access their latest GPT models\cite{OpenAI_API:online}. GPT models are trained on natural language and these models can generate responses based on their input. This input is called prompt. Through the strategic creation of tailored prompts, a diverse array of tasks can be effectively accomplished. These tasks include drafting comprehensive documents, skillfully composing computer code, conducting insightful analyses of texts, adeptly crafting conversational agents, and proficiently translating languages. Essentially, creating a prompt involves ``programming" a GPT model, which is often accomplished by providing guidelines or examples that show the model how to complete a task. 

\begin{figure*}[tbp]
\centerline{\includegraphics[width=0.87\textwidth]{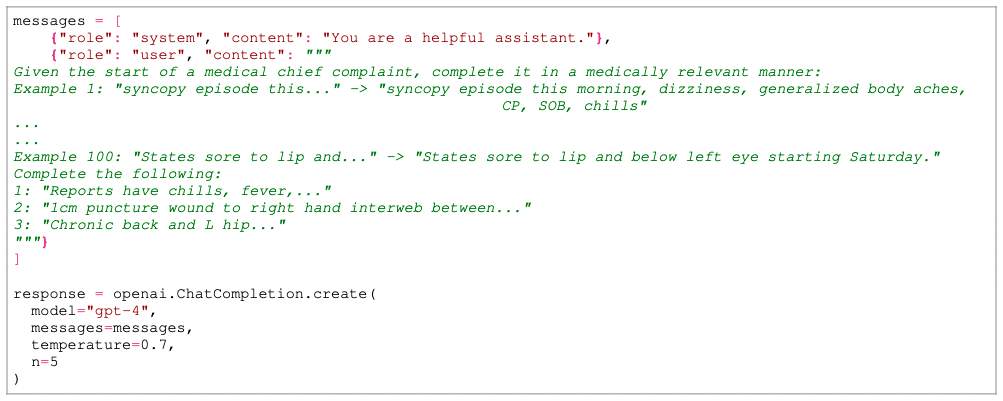}}
\caption{Prompt Tuning Code Snippet}
\label{prompt_snippet}
\end{figure*}

For our task, we tune a prompt using the OpenAI API of the GPT-4 model, which is the latest model at present. FS prompting technique is incorporated to generate CC. Although LLMs exhibit impressive zero-shot performance, they nevertheless fall short when applied to more challenging tasks. FS technique involves providing the model with a limited number of task demonstrations during the inference phase as a form of conditioning, without making any adjustments to the model\textquotesingle s weights \cite{wang2020generalizing, brown2020language}. In FS prompting technique, a handful of demonstrations are provided which lead the model towards better performance and facilitate contextual learning. According to the OpenAI official API documentation, it is recommended to have 50-100 examples as training examples, however, a minimum of 10 examples are required. \cite{OpenAI_API:online}. We chose 100 examples of varying structures from the training CC dataset for our prompt. A sample code is shown in Figure \ref{prompt_snippet}.

In the prompt development, we use OpenAI\textquotesingle s chat completions API endpoint, setting the parameter ‘temperature’ as 0.7 and ‘n’ as 5. Here ‘n’ means the number of sequences the model will generate for each input sentence. The ‘temperature’ controls the randomness of the model. Higher temperature makes the model\textquotesingle s output more diverse and random. With a higher temperature, the model may produce unusual or unexpected responses. A lower temperature makes the model\textquotesingle s output more deterministic. If the temperature is set to 0, the model will always pick the most probable next word. The outcomes are often neither overly random nor overly predictable when the temperature is moderate. 

One problem with LLM like GPT is ‘hallucination’: the creation of unreliable, irrelevant, or false information \cite{OpenAI_API:online}. GPT-4 is less likely to hallucinate than GPT-3.5-turbo. By providing explicit instructions in the prompt, it is possible to reduce hallucinations. In our proposed task, the model will suggest CC and there will be an expert in the loop. So there is minimal impact of hallucination. 

\section{Results}

The assessment of NLG model output presents considerable difficulties due to the intrinsic uncontrolled nature of many NLG tasks. In contrast to tasks with well-defined parameters that allow for definitive outputs, open-ended NLG tasks can produce a diverse array of valid and logically consistent outputs, posing challenges for objective evaluation. Consequently, conventional criteria for assessing accuracy may be inadequate, thereby requiring human judgment to evaluate the quality and relevancy of the generated content. Celikyilmaz et al. (2020) categorize the assessment approaches for NLG into three main groups: Human-Centric evaluation, Untrained Automatic Metrics, and Machine-Learned Metrics \cite{celikyilmaz2020evaluation}. In order to assess the performance of our models, we employ various methodologies such as the perplexity measure\cite{Perplexity:online}, BERTScore metric\cite{zhang2019bertscore}, and cosine similarity measure \cite{rahutomo2012semantic, farouk2019measuring}. We also include a few examples of models\textquotesingle \space output for demonstration.

\subsection{Perplexity Measure}\label{Sec:Perplexity}
Perplexity is the often employed metric for quantifying progress in language modeling \cite{jozefowicz2016exploring, dam2016deep}.
To evaluate the models\textquotesingle \space performance, we use perplexity as an evaluation metric. The metric quantifies the degree of ambiguity or perplexity exhibited by the model in its predictions of the subsequent word within a given sequence. A model\textquotesingle s performance is considered better when its perplexity score is lower, and conversely, worse when the perplexity value is higher. The concept of perplexity is characterized by the exponential value of the average negative log-likelihood of a given sequence. The perplexity of a tokenized sequence X, denoted as $X = ( x_0 , x_1 , … , x_t )$, can be calculated using Equation~\ref{Eq:Perplexity}, where $\log p_\theta\left(x_i \mid x_{<i}\right)$ denotes the $i^{th}$ tokens\textquotesingle \space log-likelihood depending on the value of preceding tokens \cite{Perplexity:online}. 

\begin{equation}
\operatorname{PPL}(X)=\exp \left\{-\frac{1}{t} \sum_i^t \log p_\theta\left(x_i \mid x_{<i}\right)\right\}
\label{Eq:Perplexity}
\end{equation}

Table \ref{result-table} provides an overview of the perplexity scores associated with our various experimented models.
From the table, we can see that the perplexity score for LSTM stands notably higher,  with an overall score of $170$. 
Hence LSTM is eliminated from further assessment. 
BioGPT, BioGPT-Large, and BioGPT-Large-PubMedQA exhibit closely aligned performance, with perplexity rates of 3.45, 1.65, and 2.20, respectively. Given the superior performance of the fine-tuned BioGPT models in comparison to the LSTM model, these three models are selected for further quantitative evaluation.

\begin{table}[tbp]
  \caption{Perplexity Score \& Execution Time}
  \label{result-table}
  \centering
  \begin{tabular}{lll}
    \toprule
    Model & Perplexity & Execution Time $^{\mathrm{a}}$ \\
    & & (milliseconds) \\
    \midrule
    LSTM & $170\pm30$ & 3727.09  \\
    BioGPT & $3.45\pm0.05$ & 9710.04 \\
    BioGPT-Large & $1.65\pm0.10$ &  30899.77 \\
    BioGPT-Large-PubMedQA & $2.20\pm0.10$ & 33584.21 \\
    \bottomrule
    \multicolumn{3}{l}{$^{\mathrm{a}}$Execution time measured on Raj-HPC\cite{Raj:online}}
  \end{tabular}
\end{table}

\begin{table}[tbp]
\centering
\caption{Comparison of BERTScore}
\label{table-BERTScore}
\begin{tabular}{*6l}
\toprule
Model & $F_{\mathrm{BERT}}$ & \multicolumn{2}{c}{All 5 CC} & \multicolumn{2}{c}{Top 2 CC}\\
\cmidrule(r){3-4}
\cmidrule(r){5-6}
{}   & {} & 30\%   & 50\%    & 30\%   & 50\%\\
\midrule
& 0.95 &  0 & 0   & 0 & 0\\
& 0.90 &  0 & 1  & 0 & 6\\
BioGPT &  0.80 & 61 & 309   & 489  & 866\\
& 0.70 &  939 &  804   & 674  & 303\\
& \textless0.70 & 177 &  63  & 14 & 2\\
\midrule
& 0.95 &  0 & 0  & 4 & 16\\
& 0.90 &  1 & 37  & 39 & 295\\
BioGPT-Large &  0.80 & 449 & 771   & 893  & 810\\
& 0.70 &  685 &  361   & 240 & 55\\
& \textless0.70 & 42 &  8  & 1 & 1\\
\midrule
& 0.95 &  0 & 0   & 1 & 19\\
& 0.90 &  2 & 27  & 45 & 291\\
BioGPT-Large- &  0.80 & 453 & 823   & 875  & 812\\
PubMedQA & 0.70 &  675 &  314   & 254  & 54\\
& \textless0.70 & 47 &  13  & 2 & 1\\
\bottomrule
\end{tabular}
\end{table}

\subsection{BERTScore Measure}\label{Sec:BERTScore}
The BERTScore measure, introduced by Zhang et al. (2019), is a recently developed method for evaluating the quality of language generation. It utilizes pre-trained BERT contextual embeddings as its foundation \cite{zhang2019bertscore}. The purpose of this system is to measure the semantic similarity between two sentences by using pairwise cosine similarity, rather than relying solely on basic string matching. In the present study, Clinical BERT \cite{alsentzer2019publicly} embeddings are employed in place of the conventional BERT embedding. The clinical BERT model has undergone pre-training on the clinical text and is accessible to the public. The procedure for computing BERTScore is implemented \cite{zhang2019bertscore}, as outlined in Equations \ref{Eq:Rbert}, \ref{Eq:pbert} and \ref{Eq:Fbert}. 
The tokenized reference sentence $x = ⟨x_1,...,x_k⟩$ is embedded into a sequence of vectors, and similarly, the tokenized candidate sentence $\hat{x} = ⟨\hat{x}_1,...,\hat{x}_l⟩$ is transformed into contextual embedding. 

\begin{equation}
    R_{\mathrm{BERT}}=\frac{1}{|x|} \sum_{x_i \in x} \max _{\hat{x}_j \in \hat{x}} \mathbf{x}_i^{\top} \hat{\mathbf{x}}_j
    \label{Eq:Rbert}
\end{equation}

\begin{equation}
P_{\mathrm{BERT}}=\frac{1}{|\hat{x}|} \sum_{\hat{x}_j \in \hat{x}} \max _{x_i \in x} \mathbf{x}_i^{\top} \hat{\mathbf{x}}_j
\label{Eq:pbert}
\end{equation}

\begin{equation}
F_{\mathrm{BERT}}=2 \frac{P_{\mathrm{BERT}} \cdot R_{\mathrm{BERT}}}{P_{\mathrm{BERT}}+R_{\mathrm{BERT}}}
\label{Eq:Fbert}
\end{equation}

Table \ref{table-BERTScore} presents the BERTScore values obtained from three BioGPT models.  We evaluate the BERTScore in 2 scenarios by selecting the seed sequence as described in section \ref{sec_data_preprocess}. For the first scenario, 30\% of each test CC is taken from the beginning as 30\% seed sequence, and for the second scenario, we take 50\%  of each test CC from the beginning as 50\% seed sequence. Each scenario is divided into 2 cases. For the first case, we consider all 5 generated CCs, and for the other case, we consider only the best 2 performing CCs. Overall we categorize our results into 4 major categories. 


\begin{itemize}
\item Scenario 1: 30\% seed sequence, All 5 generated CCs
\item Scenario 2: 50\% seed sequence, All 5 generated CCs 
\item Scenario 3: 30\% seed sequence, Top 2 generated CCs
\item Scenario 4: 50\% seed sequence, Top 2 generated CCs
\end{itemize}

For example, in Scenario 3 for BioGPT-Large, there are 39 reference test CC which achieved a BERTScore between 0.90 to 0.94. This means that there are 39 reference test CCs, whose top two generated candidate CCs achieved a BERTScore between 0.90 to 0.94, when 30\% of the reference CCs are given to BioGPT-Large as the seed sequence.

\subsection{Cosine Similarity Measure}\label{Sec:Cosine}
Table \ref{table-cosine} presents the cosine similarity score between the reference and candidate CC. In the current study, we employ the method of averaging word vectors\cite{farouk2019measuring} to compute the similarity of sentences, as denoted by the following Equation \ref{EQ:similarity}. The Clinical BERT\cite{alsentzer2019publicly} is employed to generate word embeddings. To analyze cosine similarity, we also categorize our result into 4 major categories similar to \ref{Sec:BERTScore}.

\begin{equation}
\operatorname{Similarity}(\mathbf{x}_i, \hat{\mathbf{x}}_j)=\frac{\overline{\mathbf{x}_i} \cdot \overline{\hat{\mathbf{x}}_j}}{\|\overline{\mathbf{x}_i}\| \times\|\overline{\hat{\mathbf{x}}_j}\|}
\label{EQ:similarity}
\end{equation}

\begin{table}[tbp]
\centering
\caption{Comparison of Similarity}
\label{table-cosine}
\begin{tabular}{*6l}
\toprule
Model &  Similarity & \multicolumn{2}{c}{All 5 CC} & \multicolumn{2}{c}{Top 2 CC}\\
\cmidrule(r){3-4}
\cmidrule(r){5-6}
{}   & {(Cosine)} & 30\%   & 50\%    & 30\%   & 50\%\\
\midrule
& 0.95 &  9 & 52   & 112 & 297\\
& 0.90 &  379 & 497  & 792 & 731\\
BioGPT &  0.80 & 735 & 580   & 271  & 148\\
& 0.70 &  50 &  45   & 2  & 1\\
& \textless0.70 & 4 &  3  & 0 & 0\\
\midrule
& 0.95 &  62 & 265   & 305 & 660\\
& 0.90 &  613 & 627  & 695 & 445\\
BioGPT-Large &  0.80 & 474 & 278   & 175  & 72\\
& 0.70 &  28 &  7   & 2  & 0\\
& \textless0.70 & 0 &  0  & 0 & 0\\
\midrule
& 0.95 &  55 & 248   & 291 & 644\\
& 0.90 &  606 & 643  & 685 & 472\\
BioGPT-Large-&  0.80 & 490 & 276   & 198  & 60\\
PubMedQA & 0.70 &  26 &  10   & 3  & 1\\
& \textless0.70 & 0 &  0  & 0 & 0\\
\bottomrule
\end{tabular}
\end{table}

\subsection{Execution Time Evaluation}

To evaluate the execution time of our models, we utilize the first example from Table \ref{output-table}. For each model, we generate 5 output sequences. In the context of the LSTM model, the next 5 consecutive words are predicted for each sequence. Instead of always selecting the word with the highest probability, randomness is introduced into the predictions for this model. On the other hand, BioGPT models are capable of predicting the end of the sentence. So we generate 5 full sequences with BioGPT models. As evident from Table \ref{result-table}, there is a direct correlation between model size and execution time. For instance, the LSTM model has only 1,502,676 parameters and it requires only 3727 milliseconds to generate 5 sequences. In contrast, BioGPT-Large has 1.5 billion parameters and it demands 30899 milliseconds for the same task. Broadly speaking, a model\textquotesingle s execution time is influenced by a myriad of factors, encompassing model dimensions, parameter count, architectural design, and the intricacies of the assigned task.

\begin{table*} [tb]
  \caption{Example of Generated Chief Complaints}
  \label{output-table}
  \centering
  \begin{tabular}{llp{11.5cm}}
    \toprule
    Example & Model & Candidate CC \\
    \midrule
    1  & (Reference CC) & Reports have chills, fever, \textbf{cough, CP, sore throat, back and leg pain.}\\
    \cmidrule(r){2-3}
    & BioGPT & Reports have chills, fever, \textbf{malaise x 4 days} \\
    \cmidrule(r){3-3}
    & & Reports have chills, fever, \textbf{chills, nausea, HA.} \\
    \cmidrule(r){2-3}
    & BioGPT-Large & Reports have chills, fever, \textbf{bodyaches, cough x1 week.} \\
    \cmidrule(r){3-3}
    & & Reports have chills, fever,  \textbf{dysuria, symptoms since last night.} \\
    \cmidrule(r){2-3}
    & BioGPT-Large-PubMedQA & Reports have chills, fever, \textbf{generalized malaise, diarrhea, and congestion since yesterday.}\\
    \cmidrule(r){3-3}
    & & Reports have chills, fever, \textbf{fatigue, loss of appetite.}\\
    \cmidrule(r){2-3}
    & GPT-4  & Reports have chills, fever, \textbf{fatigue, and sore throat x 4 days, tested negative for Covid-19}  \\
    \cmidrule(r){3-3}
    & (Prompt) & Reports have chills, fever, 
    \textbf{coughing and headaches for the past 3 days} \\
    \hline
    2  & (Reference CC) & 1cm puncture wound to right hand interweb between \textbf{2nd and 3rd digit, tetanus UTD, denies pmh}\\
    \cmidrule(r){2-3}
    & BioGPT & 1cm puncture wound to right hand interweb between \textbf{thumb and hand.} \\
    \cmidrule(r){3-3}
    & & 1cm puncture wound to right hand interweb between \textbf{wound to R middle finger.} \\
    \cmidrule(r){2-3}
    & BioGPT-Large &  1cm puncture wound to right hand interweb between \textbf{2nd and 3rd digit, tetanus, denies PMH} \\
    \cmidrule(r){3-3}
    & & 1cm puncture wound to right hand interweb between \textbf{2nd and 3rd finger} \\
    \cmidrule(r){2-3}
    & BioGPT-Large-PubMedQA & 1cm puncture wound to right hand interweb between \textbf{2nd and 3rd digit, swelling and pain to wound.} \\
    \cmidrule(r){3-3}
    & & 1cm puncture wound to right hand interweb between \textbf{2nd and 3rd digit, tetanus not UTD} \\
    \cmidrule(r){2-3}
    & GPT-4  & 1cm puncture wound to right hand interweb between \textbf{thumb and index finger, no signs of infection but pain is increasing.} \\
    \cmidrule(r){3-3}
    & (Prompt) & 1cm puncture wound to right hand interweb between \textbf{thumb and index finger, caused by a rusty nail.} \\
    \hline
    3  & (Reference CC) & Chronic back and L hip \textbf{pain x ``years" and R shoulder pain x 1 month.}\\
    \cmidrule(r){2-3}
    & BioGPT & Chronic back and L hip \textbf{pain x 2 years, denies pmh} \\
    \cmidrule(r){3-3}
    & & Chronic back and L hip \textbf{pain, worse with ambulation x one week} \\
    \cmidrule(r){2-3}
    & BioGPT-Large &  Chronic back and L hip  \textbf{pain x1 year.} \\
    \cmidrule(r){3-3}
    & & Chronic back and L hip \textbf{pain x1 week.} \\
    \cmidrule(r){2-3}
    & BioGPT-Large-PubMedQA & Chronic back and L hip \textbf{pain, denies trauma, no known falls} \\
    \cmidrule(r){3-3}
    & & Chronic back and L hip \textbf{pain, radiating down R leg x1 year.} \\
    \cmidrule(r){2-3}
    & GPT-4  & Chronic back and L hip \textbf{pain, exacerbated by movement, no relief with OTC pain medication.} \\
    \cmidrule(r){3-3}
    & (Prompt) & Chronic back and L hip\textbf{ pain, worsening over last week, OTC meds provide no relief.} \\
    \bottomrule
    \multicolumn{3}{l}{*No objective
metric is reported in Table~\ref{table-BERTScore} and \ref{table-cosine} for GPT-4 prompt tuning output.}
  \end{tabular}
\end{table*}

\section{Discussions}
The language structures seen in clinical documentation are complex and diverse as a result of the specific nature of medical information and terminologies. In addition, the acquisition of clinical text datasets poses a persistent challenge due to the ethical considerations around patient privacy and the unique nature of medical narratives. In our study, we found that there is a correlation between the size of a corpus and the perplexity score of a Language Model. Larger corpora tend to yield higher scores, indicating improved performance \cite{jozefowicz2016exploring, kauchak2013improving}. Deep learning models tend to get advantages from an increased quantity of training data. Typically, the efficacy of training an LSTM model relies upon the availability of a substantial volume of data, particularly for tasks of greater complexity. This is because the model needs to learn more nuanced patterns in the data to make accurate predictions. Insufficient information within a short dataset may impede the model\textquotesingle s ability to properly learn, resulting in inferior outcomes. The performance of our baseline LSTM model is suboptimal, mostly attributed to the limited size of our corpus. 

Based on the perplexity score presented in Table \ref{result-table}, it can be observed that large BioGPT models exhibit a higher level of performance compared to BioGPT.  Tables \ref{table-BERTScore} and \ref{table-cosine} also demonstrate similar findings. In every scenario, large models consistently outperform BioGPT in terms of scoring. In Table \ref{table-BERTScore} Scenario 1, large models display approximately 450 reference test CCs, exceeding a BERTScore of 0.80. On the other hand, the BioGPT model manages only 61 reference test CCs. For Scenario 2, around 70\% of the reference test CCs for large models reach a BERTScore of 0.80 or above, whereas BioGPT shows results for less than 30\% of the reference test CCs. In Scenario 3, more than 80\% of the reference test CCs for large models hit a BERTScore of 0.80 or more. Lastly, in Scenario 4, the large models are excellent, with almost all reference test CCs reaching a BERTScore of 0.80 or above.

When we select 50\% seed sequence instead of 30\%, all our models achieve superior BERTScore. One of the plausible reasons behind this is that it becomes easier to generate the incomplete portion when more clues are given. Among all of the scenarios considered for large models, it can be observed that BERTScore performs less well in Scenario 1. Given that we are taking into account all five candidate CCs that have been generated, it is also important to note that only 30\% of the test reference CC is being utilized as input for the models. On the other hand, the models have exhibited exceptional performance in Scenario 4. This can be attributed to the fact that we have only focused on the top two performing candidate CCs, with 50\% seed sequence as input.

According to the data shown in Table~\ref{table-cosine}, while utilizing the semantic cosine similarity measure, it is observed that large models achieve a similarity score of 0.90 for 60\% reference CC in Scenario 1, and around 95\% reference CC in Scenario 4. BioGPT models especially large models show promising performance in generating contextually similar CCs. 

In table \ref{output-table}, for demonstration we provide a few examples of models\textquotesingle \space output including GPT-4 prompt tuning. No objective metric is reported for prompt tuning. In the table, reference CC is shown in the first row of every example. The models generate the bold-face part and the first part of the reference CC is given to the models as seed sequence. 
In example 1, the patient reports several symptoms such as chills, fever, etc. Our BioGPT-Large model is able to generate a few related symptoms such as bodyaches, cough, etc. The model not only suggests related symptoms but also proposes a time. The recommendation of time will help triage nurses improve their clinical notes. 
BioGPT predicts a few irrelevant symptoms such as `chills' which are already present in the sentence. BioGPT-Large-PubMedQA generates some relevant symptoms and a probable timeframe, which is quite similar to the output of BioGPT-Large model. 
In example 2, when 50\% seed sequence is given, both BioGPT-Large and BioGPT-Large-PubMedQA are able to complete the phrase and suggest the next words almost similar to reference CC. However, BioGPT fails to generate a meaningful CC sentence in this scenario. 
In example 3, the reference CC has 2 parts formed as a compound CC. Each of our experimented models successfully predicts the next word `pain'. Though the BioGPT-Large model was able to complete the phrase, it failed to generate the last part. Other models could not capture the first phrase properly. Several CCs consist of multiple clauses and also include direct statements made by patients.  Such a CC is - about 7wks pregnant per pt, pt thinks she\textquotesingle s having a miscarriage, pt states, “last night I felt like I was bleeding more than spotting”. The performance of our experimented models for these particular sorts of CC is comparatively inferior. 

For all of these 3 aforementioned examples, GPT-4 successfully generates meaningful long sentences. However, from our observation, it seems unable to capture the CC structure fully. Overall, our fine-tuned BioGPT-Large model performs better. 

Though our fine-tuned BioGPT-Large model works excellently in the short term, it diverges in the long term. It\textquotesingle s not uncommon for language models like BioGPT to perform well in generating short-term text, but struggle with generating longer sequences. This is because generating long sequences requires the model to maintain coherence and consistency over a larger context, which can be challenging even for SOTA models. 
In the training set, the median number of words in a CC sentence is 9. It is expected that user input will be 3 or 4 words which is 30\% to 50\% of the CC sentence. As a result, suggesting the next 5 subsequent words will prevent divergence. If 5 words are not required to complete a sentence, the BioGPT-Large model holds the capability to predict the end of a sentence; exhibit example 3 in Table \ref{output-table}.

\section{Conclusion and Future work}

To conclude, we evaluate the performance of two different types of language models, LSTM and BioGPT, for generating CCs. Our results show that the BioGPT models outperform the LSTM model in terms of perplexity score. We further evaluate BioGPT models based on BERTScore and cosine similarity. Among all BioGPT models, BioGPT-Large achieves superior performance while generating more accurate and coherent CC. In addition, we identify that the performance of the LSTM model is limited due to the small size of our training data. 

In the upcoming phase, we intend to conduct a Human-Centric evaluation of our models\textquotesingle \space outputs, with insights from domain experts. Additionally, we will use a medical corpus to ensure the accuracy of medical terminologies. Moreover, we aim to refine the date-time representation during post-processing.

\section*{Acknowledgment}
We extend our sincere gratitude to Dr. Nasim Yahyasoltani and Kevin Chovanec from MU, as well as Ahnaf Farhan from UTEP, for their invaluable suggestions during this work.

\bibliographystyle{IEEEtran}
\bibliography{ref}
\vspace{12pt}
\end{document}